\documentclass[11pt]{article}

\usepackage[preprint]{acl}

\usepackage{times}
\usepackage{latexsym}
\usepackage{listings}
\usepackage{graphicx}
\usepackage{tabularx}
\usepackage{subcaption}

\usepackage[T1]{fontenc}

\usepackage[utf8]{inputenc}

\usepackage{microtype}

\usepackage{inconsolata}

\usepackage{amsmath} 

\usepackage{booktabs}

\usepackage{tikz, pgfplots}
\pgfplotsset{compat=1.18} %
\usetikzlibrary{shapes.geometric, positioning, shadows, arrows.meta}

\newcommand{\rmspace}{\vspace{-2ex}}

\title{The Right Information Extraction Pipeline Depends on the Document:\\
Accuracy--Energy Trade-offs for Small, Local Models}

\author{
  Christoph Walser \\
  Zurich University of \\
  Applied Sciences, Switzerland \\
  \small\texttt{walchr01@students.zhaw.ch}
  \And
  Mauricio Fadel Argerich \\
  Universidad Politécnica \\
  de Madrid, Spain \\
  \small\texttt{mauricio.fadel@alumnos.upm.es}
  \And
  Jonathan Fürst \\
  Zurich University of \\
  Applied Sciences, Switzerland \\
  \small\texttt{jonathan.fuerst@zhaw.ch}
}

\begin{document}
\maketitle
\begin{abstract}
Whether an information extraction pipeline should process page images or parsed text depends on the document, and the answer flips across the layout spectrum. We study this trade-off under a constraint that rules out (closed) cloud services: privacy-sensitive documents processed on-premise by small ($\le 8\mathrm{B}$ parameter) text-only and vision--language models, evaluated on both accuracy and energy over a design space spanning input representation, model family, and inference configuration. Benchmarking on the near-plain-text Kleister-NDA contracts and the layout-rich VRDU forms, we find that batching is the dominant energy lever, cutting energy per page by 38--85\% at no cost in accuracy, while FP8 quantization saves 27--32\% when requests are served one at a time but less than 1\,mWh per page (9--19\%) once batching is applied. Preprocessing dominates what remains: neural OCR costs $17\times$ more energy per page than classical OCR and never reaches the Pareto frontier. Which representation wins flips with the type of document: vision--language models on layout-rich documents and small text-only models with a cheap parser on near-plain text, where they are both more accurate and cheaper than any vision--language configuration. Our work yields concrete guidelines for energy-efficient, privacy-compliant local information extraction.
\end{abstract}

\section{Introduction}

Documents in practice span a wide layout spectrum. At one end sit layout-rich documents (e.g., registration forms, invoices), where the meaning of a value depends on its position in a two-dimensional arrangement of boxes and tables. At the other sit  near-plain text documents, such as contracts and reports,  that encode almost nothing in their visual structure \cite{ding2024survey, watanabe1995layout}. Extraction pipelines are not indifferent to this: a model that reads page images directly has an advantage where layout carries meaning, and pays a cost for a modality it does not need when processing near-plain text documents.

Large Language Models (LLMs) and Vision Language Models (VLMs) have transformed key information extraction (KIE), replacing rigid rule-based heuristics~\cite{luo2024layoutllm, colakoglu2025problem}. However, their deployment remains constrained: Large transformer architectures demand computational resources that resource-constrained settings cannot supply~\cite{caravaca2026}, and routing sensitive documents to third-party providers risks data leakage and frequently violates data protection regulations in domains like finance and healthcare~\cite{ding2024survey}. Both pressures push toward small, locally deployable models running on-premise under tighter data governance. We therefore propose KIE should be considered \textit{not as a prediction problem alone but as a multi-objective one, balancing extraction quality against end-to-end energy}, across small ($\leq 8\mathrm{B}$ parameter) general-purpose LLMs, VLMs, and specialized architectures.

Following this proposal, we benchmark pipelines at both ends of the layout spectrum: the near-plain-text Kleister-NDA contracts~\cite{kleister-nda} and the densely structured VRDU registration forms~\cite{vrdu-dataset}. The design space covers visual-first processing on raw page images and text-first processing via OCR, across model families and inference configurations. Holding the model, parser and serving configuration fixed across two corpora that differ in both layout and length is what lets us show that the preferred pipeline inverts between them. The main contributions of our work are:
\begin{itemize}
    \item A systematic comparison of local document IE pipelines from an accuracy--energy perspective.
    \item An analysis of how preprocessing choices (native vision vs. OCR vs. parser) impact both total pipeline energy and downstream extraction accuracy.
    \item An evaluation of deployment optimizations, showing that request batching reduces energy per page by 38--85\% without harming extraction quality, and that the benefit of FP8 quantization shrinks roughly tenfold once batching is in place.
    \item Practical deployment recommendations for building accurate, energy-efficient, and privacy-compliant IE pipelines.
\end{itemize}

All configurations and code to reproduce our results are available at \url{https://github.com/chrewbroccoli/local-ie-energy}. 

\section{Related Work}
A growing body of work measures the energy of large language model (LLM) inference and proposes strategies for sustainable deployment~\cite{luccioni2024power,fernandez2025energy,poddar2025benchmarking,argerich2024measuring}. \citet{luccioni2024power} show that inference, not training, dominates the lifetime energy cost of widely deployed models, and that cost varies by orders of magnitude across task types. \citet{fernandez2025energy} show that optimizations such as batch size, decoding strategy, and serving stack (e.g., PyTorch, CUDA Graphs, vLLM) are highly sensitive to workload geometry and hardware, and that appropriate configurations can cut total energy use by up to 73\%. Complementary benchmarks measure energy per request across models and concurrency levels, finding that cost per request falls as batch size and concurrency increase \cite{poddar2025benchmarking,pronk2025benchmarkingenergyefficiencylarge}. Watt Counts~\cite{argerich2026wattcountsenergyawarebenchmark} extends such measurements to 50 LLMs on 10 GPU architectures and shows that the most energy-efficient GPU differs across models and deployment scenarios, while WattGPU~\cite{argerich2026wattgpupredictinginferencepower} predicts inference power and latency on unseen GPUs and LLMs from public specifications alone. Bench360~\cite{stuhlmann2026bench360benchmarkinglocalllm} provides a framework for profiling local LLM inference with custom tasks.
 \citet{argerich2024measuring} advocate task-agnostic guidelines: select efficient models, use quantized variants where available, and increase batch size until GPU utilization saturates. \citet{delavande2026understanding} qualify this, showing that lower-precision formats gain most in compute-bound prefill phases while dequantization overheads can negate the savings, and that batching improves efficiency across all inference phases by reducing GPU idle time. These findings motivate our exploration of small, locally deployable models, FP8 quantization, and vLLM-backed batch inference.

LLM-based key information extraction (KIE) has advanced from classic text-only IE to layout-aware and multimodal document understanding~\cite{xu2024llm-ie,ding2024survey}. Surveys of generative IE summarize prompting and instruction-tuning strategies and show that general-purpose LLMs can match specialized architectures on many IE tasks~\cite{xu2024llm-ie}. LayoutLLM introduces layout instruction tuning so that LLMs can process serialized two-dimensional document structures~\cite{luo2024layoutllm}, building on earlier work that combines text, layout, and visual cues for extraction from complex business documents~\cite{katti2018chargrid,denk2019bertgrid,xu2020layoutlm,borchmann2021due,borchmann2025arctictilt, kleister-nda,vrdu-dataset}. A parallel line removes the text-extraction stage entirely: OCR-free architectures such as Donut~\cite{kim2022donut} and Pix2Struct~\cite{lee2023pix2struct} read page images directly, a strategy today's general-purpose VLMs inherit and that we evaluate here from an energy perspective. At the complex end of the spectrum, agentic pipelines built on large cloud models plan and decompose extraction from dense, multi-page, multilingual regulatory documents~\cite{colakoglu2026agenticieadaptiveagentinformation}. We bridge these lines by measuring accuracy and end-to-end energy across the layout spectrum rather than at a single point on it, spanning OCR technologies, small text-only and multimodal LLMs, specialized layout models, and batch- and quantization-aware vLLM inference.

\section{Design Space of Resource-Aware IE with Small Local LLMs}
Figure~\ref{fig:design_space} summarizes the design space: for a given document and schema we need to decide the input representation, the model and the serving configuration under energy and accuracy constraints.

\newcommand{\Card}[2]{
    \begin{minipage}{\linewidth} \raggedright \sffamily
        {\centering \textbf{#1} \par}   %
        \vspace{0.15cm}                 %
        #2                              %
    \end{minipage}%
}
\newcommand{\SmallCard}[2]{%
    \begin{minipage}{\linewidth} \raggedright \sffamily
        {\centering \textbf{#1} \par}   %
        \vspace{0.15cm}                 %
        #2                              %
    \end{minipage}%
}

\newcommand{\entry}[1]{\par {\small \textbullet~#1} \vspace{2pt}}

\begin{figure}[t!]
    \centering
    \resizebox{\columnwidth}{!}{%
    \begin{tikzpicture}[
        >={Stealth[scale=1.1]},
        DS/.style={
            rectangle, 
            draw=blue!60!black, 
            thick, 
            fill=blue!5,
            text width=2.7cm,    %
            align=left, 
            rounded corners=4pt,
            inner sep=6pt,
            minimum height=2.8cm,
            drop shadow={opacity=0.15, shadow xshift=2pt, shadow yshift=-2pt}
        },
        CYL/.style={
            cylinder,
            shape border rotate=90,
            aspect=0.15,            
            draw=gray!80!black, 
            thick,
            fill=gray!5,
            text width=2.5cm,    %
            align=left,
            inner sep=6pt,
            minimum height=2.8cm,
            drop shadow={opacity=0.15, shadow xshift=2pt, shadow yshift=-2pt}
        },
        REP/.style={
            rectangle, 
            draw=green!60!black, 
            thick, 
            fill=green!5,
            text width=2.7cm,    %
            align=left, 
            rounded corners=4pt,
            inner sep=6pt,        
            minimum height=2.8cm,
            drop shadow={opacity=0.15, shadow xshift=2pt, shadow yshift=-2pt}
        },
        node distance=0.3cm %
    ]

    \node[CYL] (Docs) {
        \Card{Dataset ($D,S$)}
        {
            \entry{Layout-rich / plain-text docs}
            \entry{Target Schema}
        }
    };
    
    \node[DS, right=of Docs] (InputRep) {
        \Card{Input Rep. ($I$)}
        {
            \entry{Raw Images}
            \entry{OCR Text-Extraction}
            \entry{Model based extraction}
        }
    };
    
    \node[DS, right=of InputRep] (Model) {
        \Card{Models ($M$)}
        {
            \entry{VLMs}
            \entry{LLMs}
            \entry{Specialized Models}
        }
    };
    
    \node[DS, right=of Model] (Quant) {
        \Card{Inference Config.($C$)}
        {
            \entry{vLLM}
            \entry{Batch / Single}
            \entry{Quantization}
        }
    };
    
    \node[REP, right=of Quant] (Results) {
        \SmallCard{Results ($E \& R$)}
        {
            \entry{Structured Data ($E$)}
            \entry{Metrics ($R$)}
        }
    };
    
    \tikzstyle{arrow style}=[->, thick, draw=gray!80!black]
    
    \draw[arrow style] (Docs.east) -- (InputRep.west);
    \draw[arrow style] (InputRep.east) -- (Model.west);
    \draw[arrow style] (Model.east) -- (Quant.west);
    \draw[arrow style] (Quant.east) -- (Results.west);
    
    \end{tikzpicture}%
    } %
    \caption{Overview of the benchmarking design space. We investigate the dimensions ($D, S, I, M, C$) to evaluate the trade-offs between extracted structured information ($E$) and resource consumption ($R$).}
\label{fig:design_space}
\end{figure}
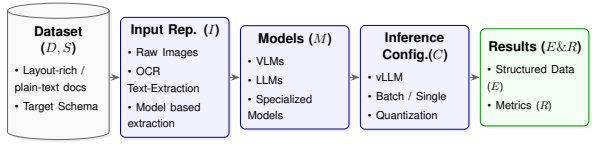

\subsection{Task Definition}
We define the resource-aware KIE task as a mapping from a document and a system configuration to a joint output space of structured data and computational cost:
\[
\text{KIE}_{\text{res}} : (D, S, I, M, C) \rightarrow (E, R)
\]
On the input side, $D$ is the document set and $S$ the target schema of fields to populate; $I$ is the input representation, either raw page images or the output of a text-extraction pipeline; $M$ is the model, whether text-only LLM, VLM, or specialized layout model; and $C$ is the inference configuration, here batch size and numeric precision. On the output side, $E$ is the extracted structured data, validated against $S$, and $R$ is the measured resource cost, here energy per page.

\subsection{Dimensions of the Design Space}

The three dimensions interact. \textbf{Input representation} ($I$) determines how a document is exposed to the model: layout is preserved implicitly through pixels (visual-first) or explicitly through serialized, layout-aware text from a parser or OCR engine (text-first). \textbf{Model architecture} ($M$) determines which information pathways are available: a VLM consumes page images directly, a text-only LLM depends entirely on the upstream parser, and a specialized layout model additionally requires word-level bounding boxes. \textbf{Inference configuration} ($C$) determines how the serving engine allocates memory and schedules requests, fixing the achievable hardware utilization. These choices are not independent: bypassing the parser removes an energy term but shifts cost into the model, and the batch size that saturates a 2B VLM differs from the one that saturates an 8B text model.

Each configuration is scored on a dual objective: extracted structures are validated against the target schema for accuracy ($E$), and the full execution is profiled for energy ($R$). We measure the entire pipeline, so parsing energy is charged to the configurations that incur it.

\section{Methodology \& Experimental Setup}
The evaluation proceeds in three phases, one per dimension of the design space, each addressing one research question.

\noindent\textbf{RQ1 Baselines.} \textit{How do small VLMs, text-only LLMs with basic OCR, and specialized models differ in their accuracy--energy profiles on text-heavy and layout-rich documents under a simple, unoptimized serving configuration?} We evaluate the three model categories using single-request inference: vision-language models processing raw images directly, text-only LLMs using Tesseract OCR as the baseline text input, and specialized layout models (Arctic-TILT and NuExtract).

\noindent\textbf{RQ2 Input Representation.} \textit{How do different document parsing strategies --- embedded-text parsers, classical OCR, and neural OCR --- affect end-to-end energy and extraction accuracy, and how does this depend on whether documents are born-digital or scanned and layout-rich?} For the text-only models we substitute the Tesseract baseline with Docling and DeepSeek-OCR 2, measuring the trade-off between text-extraction quality and the energy overhead of each mechanism.

\noindent\textbf{RQ3 Inference Optimization.} \textit{How do batch size and FP8 quantization shape the accuracy--energy trade-off?} We sweep batch size and precision under vLLM across all pipelines and measure the effect on energy per page and on extraction quality.

\subsection{Models}
We evaluate three model categories: vision language models (VLMs), text-only LLMs, and specialized layout models. Table~\ref{tab:model_categories_single} lists them with their modality and evaluation mode.

\begin{table}[ht]
  \caption{Evaluated models. Only Arctic-TILT saw a target dataset during training; all others are zero-shot.}
  \label{tab:model_categories_single}
  \centering
  \footnotesize
  \begin{tabularx}{\columnwidth}{@{} >{\raggedright\arraybackslash}X l l @{}}
    \toprule
    \textbf{Model} & \textbf{Modality} & \textbf{Eval. Mode} \\
    \midrule
    
    \multicolumn{3}{@{}l}{\textit{\textbf{Specialized Models}}} \\
    \addlinespace[2pt]
    Arctic-TILT \cite{borchmann2025arctictilt} & OCR+Layout & In-domain \\
    NuExtract-2.0-4B \cite{nuextract2} & VL & Zero-shot \\

    \midrule
    \addlinespace[2pt]
    \multicolumn{3}{@{}l}{\textit{\textbf{General-Purpose Models}}} \\
    \addlinespace[2pt]
    Qwen3-VL-Instruct \newline (2B, 4B, 8B) \cite{qwen3vl} & VL & Zero-shot \\
    \addlinespace[4pt]
    Llama-3.2-Instruct (1B, 3B) \cite{llama3} \newline
    Ministral-3-3B \cite{ministral3} \newline
    Mistral-7B \cite{mistral7b} \newline
    Qwen3 (0.6B, 1.7B, 4B, 8B) \cite{qwen3} & Text & Zero-shot \\

    \bottomrule
  \end{tabularx}
\end{table}

\subsection{Datasets}
\label{sec:datasets}
\begin{figure*}[t]
    \centering
    \begin{subfigure}[b]{0.98\columnwidth}
        \centering
        \fbox{\includegraphics[page=1, width=1.0\linewidth]{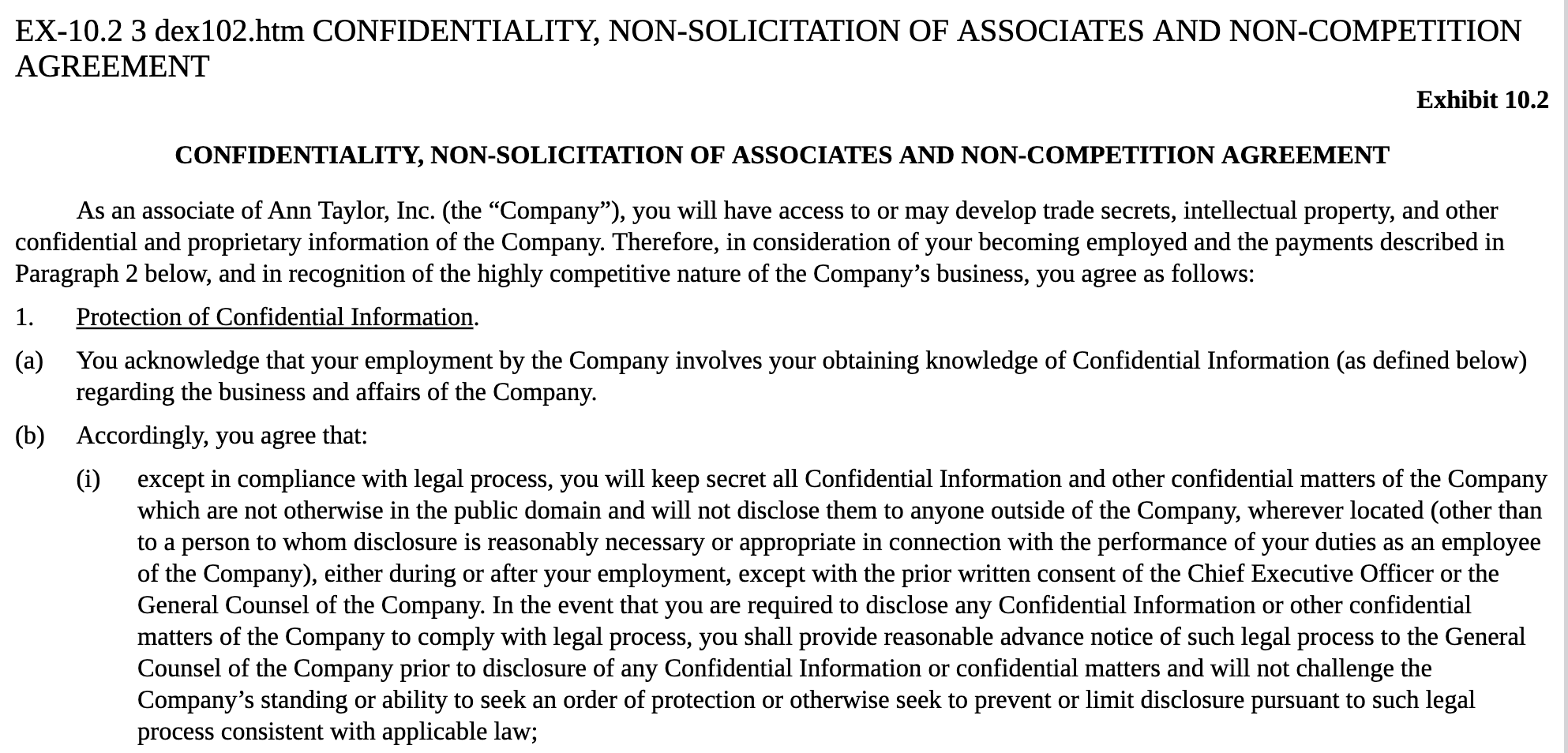}}
        \caption{Example of Document from Kleister-NDA Dataset}
        \label{fig:layout_kleister}
    \end{subfigure}
    \hfill
    \begin{subfigure}[b]{0.75\columnwidth}
        \centering
        \fbox{\includegraphics[page=1, width=1.0\linewidth]{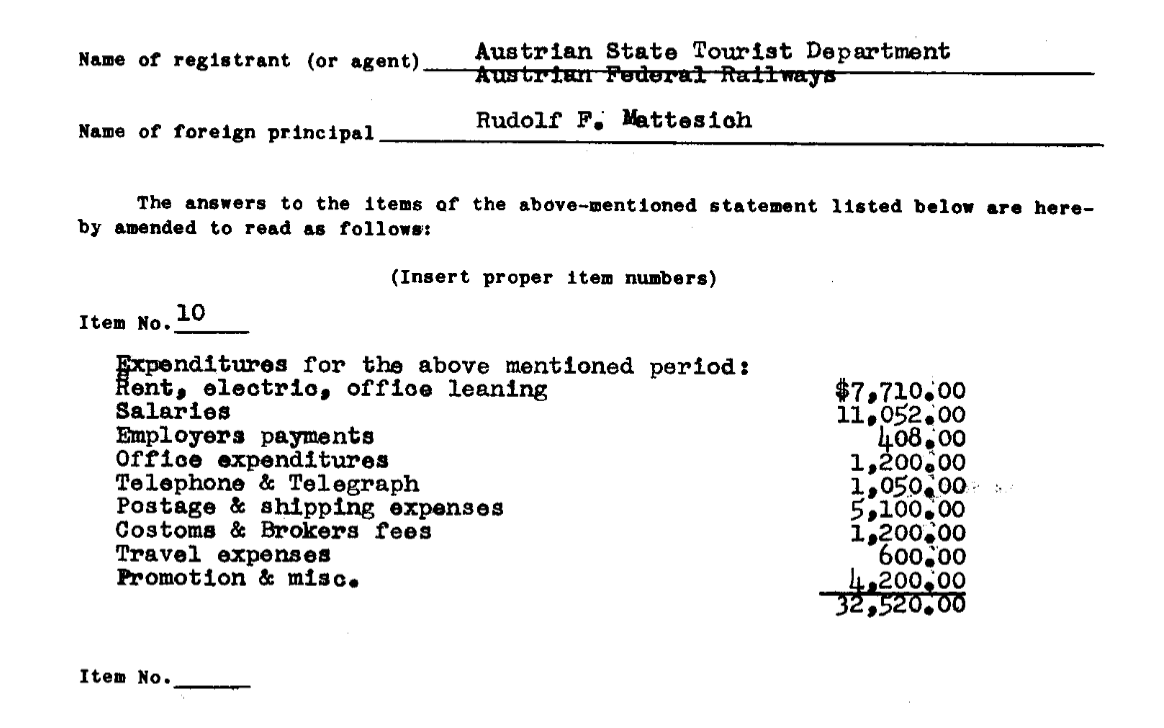}}
        \caption{Document from the VRDU dataset}
        \label{fig:layout_vrdu}
    \end{subfigure}
    
    \caption{A visual comparison between a low-layout document (from the Kleister-NDA dataset) and a highly structured document (from the VRDU dataset). Models must leverage 2D spatial features to successfully extract information from the document on the right.}
    \label{fig:document_layout_comparison}
    \rmspace
\end{figure*}

We evaluate on two datasets: the text-heavy Kleister-NDA \cite{kleister-nda} and the visually complex VRDU Registration \cite{vrdu-dataset}, with an example of each in Figure~\ref{fig:document_layout_comparison}. Information in VRDU is carried by 2D spatial arrangements and visual elements such as tables, whereas Kleister-NDA documents have minimal visual structure and follow a one-dimensional textual flow. VRDU ships its own baseline OCR text, which we use as an additional reference point when ranking parsers.

Extraction is evaluated on 500 VRDU Registration documents (up to six schema fields per document) and on 337 Kleister-NDA documents (up to four fields: \texttt{effective\_date}, \texttt{jurisdiction}, \texttt{party}, \texttt{term}). In both cases only fields present in the ground truth of a given document are requested. Parsing energy (Section~\ref{sec:parsing}) was profiled over 500 documents per dataset and combined with inference energy before normalization.

The two corpora also differ in length, which determines how we report energy. Over the same 500 profiled documents, Kleister-NDA runs to 2934 pages against 915 for VRDU, i.e.\ 5.87 against 1.83 pages per document. Energy per document would therefore conflate the cost of processing a page with the number of pages a document happens to have, and the two datasets could not be compared on it. We report energy \emph{per page} throughout.

Length is also a confound in its own right: the low-layout corpus is the long one, so layout complexity and document length co-vary between our two datasets and cannot be separated by comparing them. Normalizing by page removes length from the energy axis but not from the experiment, and vision runs additionally see at most the first 10 (Kleister-NDA) or 5 (VRDU) pages of a document. We therefore read the results as a contrast between two \emph{document types} --- short, layout-rich forms and long, near-plain-text contracts --- rather than as an isolated effect of layout; Section~\ref{sec:pareto} examines what the per-page energy data can and cannot say about length.

One asymmetry needs to be stated up front. Arctic-TILT is released already fine-tuned on Kleister-NDA~\cite{borchmann2025arctictilt}, so on that dataset it is the only model in our study that has seen the target corpus during supervised training, and part of our evaluation set is likely to overlap with its training data. This is a property of the released checkpoint rather than of our protocol, and we do not attempt to correct for it. We report Arctic-TILT's Kleister-NDA results for reference, mark them as in-domain in every table and figure, and exclude them whenever we identify a best configuration or compute a Pareto frontier on that dataset. On VRDU, which Arctic-TILT has not been trained on, it is treated like every other model.

\subsection{Document Parsing}
\label{sec:parsing}
We compare three parsers: Tesseract \cite{smith2007tesseract} (CPU), Docling \cite{Livathinosetal2025} (CPU + GPU acceleration), and DeepSeek-OCR 2 \cite{deepseekocr2} (GPU). We profile the PDF-to-Markdown conversions with CodeCarbon \cite{benoit_courty_2024_11171501}, using multi-processing for the conversion scripts and batched inference for the GPU-accelerated parser. Each parser was run over 500 documents per dataset. So that the idle power of the L4 is not charged to a CPU-bound pipeline, GPU tracking is disabled in CodeCarbon for Tesseract (\texttt{gpu\_ids=[]}), leaving only its CPU and RAM energy; Docling and DeepSeek-OCR~2 run on the GPU and are charged for it. Arctic-TILT additionally requires word-level bounding boxes, which DeepSeek-OCR 2 does not provide; we interpolate them from its paragraph-level output (Appendix~\ref{sec:preprocessing}). The parsers do not agree on page segmentation, so per-page figures are computed against each tool's own page count (Table~\ref{tab:energy_consumption_integrated}).

\subsection{Inference}
\label{sec:inference}
Inference for both text-only and vision-language models ran on a single NVIDIA L4 GPU using vLLM~\cite{kwon2023vllm}. Because the Chat Completions API is request-oriented rather than a native offline batching interface, we issue concurrent asynchronous requests and let vLLM batch them internally.
Quantization is common in resource-constrained deployments, so we evaluate each model at FP16 and FP8.
Since batch size improves resource utilization and reduces energy~\cite{argerich2024measuring}, we evaluate batch sizes of 1, 5, 10, 20, 40 and 60, subject to memory availability.
Decoding follows each model family's recommended sampling defaults, with reasoning traces disabled for the Qwen3 models; the full parameters, and evidence that no run was limited by its token budget, are given in Appendix~\ref{app:runconfig}. Because the text-only and vision arms therefore use different sampling settings, we bounded the effect by re-running the eight text models on VRDU under the vision arm's settings with everything else fixed: at batch size 10, exact match moved by $-1.6$ to $+3.2$ points (mean $+2.0$), well below the gaps of 7 points or more between size-matched text and vision models on which the modality comparison rests. Prompt templates for both datasets are listed in Appendix~\ref{app:prompt_templates}.
We benchmark the general-purpose LLMs and VLMs with Bench360 \cite{stuhlmann2026bench360benchmarkinglocalllm}, and the specialized models (NuExtract and Arctic-TILT) with custom inference scripts profiled by CodeCarbon.

\subsection{Metrics}
We report average field exact match and energy per page; F1 and fuzzy match are also computed and reported in Appendix~\ref{detailed-results}.

\noindent\textbf{Average Field Exact Match (EM):} For each document we compare the predicted and gold value sets of every requested field and report the proportion of answered fields whose set matches the gold set exactly. Values are lowercased and standardized (e.g., converting text-based numbers to digits, normalizing dates) prior to comparison. Fields that a model declines to answer are not counted in the denominator, so EM behaves as a field-level precision; the value-level F1 reported in Appendix~\ref{detailed-results} penalizes omissions and should be consulted alongside it.

\noindent\textbf{Energy Consumption:} Total energy of the execution pipeline, covering both parsing (CodeCarbon~\cite{benoit_courty_2024_11171501}) and model inference (Bench360), in milliwatt-hours per page (mWh/pg), dividing by the PDF page count of each corpus (5.87 pages per document on Kleister-NDA, 1.83 on VRDU). Table~\ref{tab:energy_consumption_integrated} is the one exception, for the reason given in Section~\ref{sec:parsing}. Both tools sample GPU power through NVML at 10\,Hz and model the CPU and RAM contributions rather than measuring them; the Limitations discuss what this implies for CPU-only pipelines. Model loading is excluded, so the figures describe steady-state serving rather than cold start.

\section{Results}

\subsection{RQ1 Baselines}
\label{sec:rq1}
\begin{figure*} [ht]
    \centering
    \includegraphics[width=\linewidth]{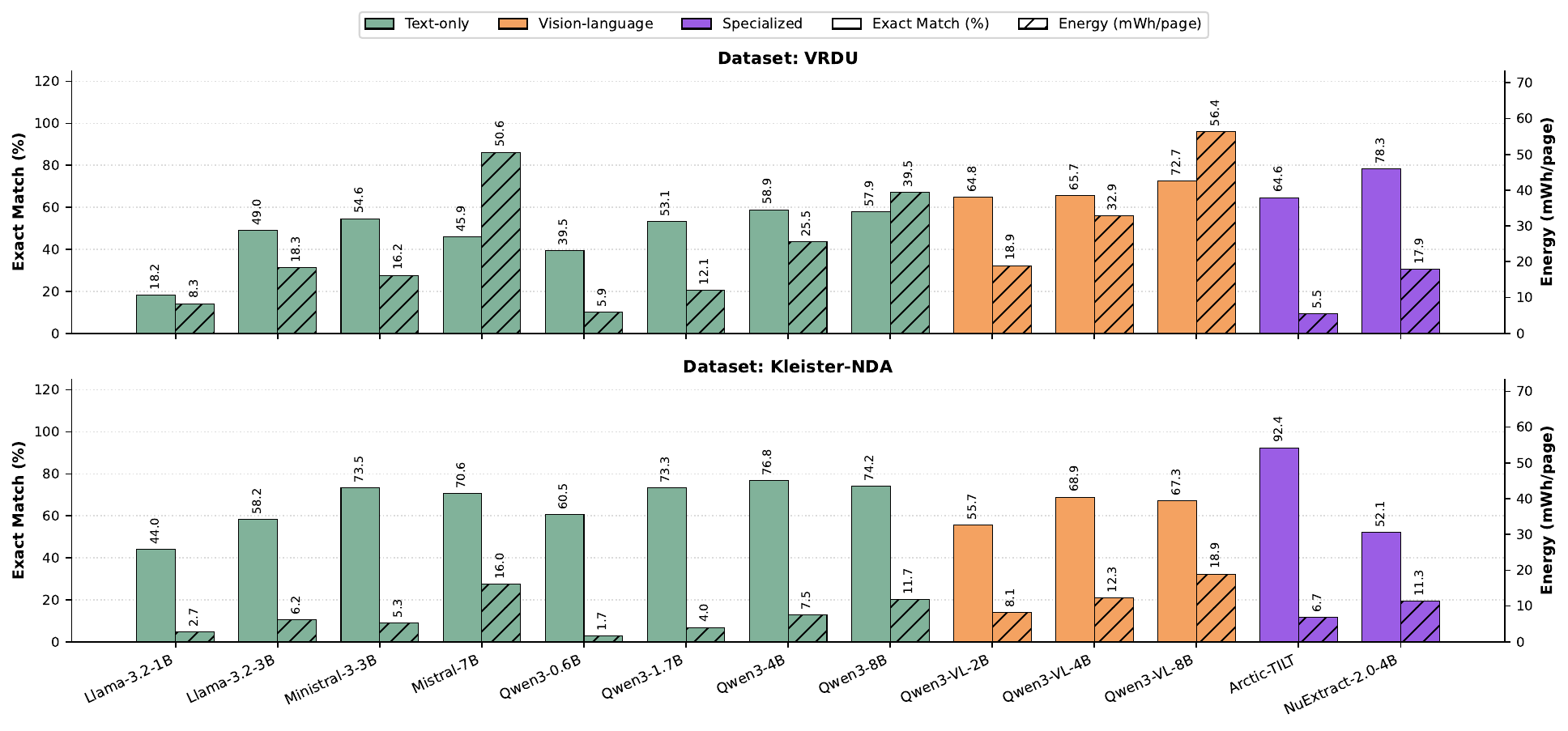}
    \caption{Baseline extraction accuracy and energy consumption per page on VRDU (top) and Kleister-NDA (bottom), at batch size 1 without quantization, using Tesseract-based PDF-to-Markdown conversion for all text-only models. Both panels share the model axis and the energy scale, so the two datasets can be read against each other. Energy covers model inference only; parsing energy is added in the end-to-end analysis of Section~\ref{sec:pareto}. Arctic-TILT's Kleister-NDA result is in-domain (fine-tuned on that corpus), shown for reference only and not part of any zero-shot comparison.}
    \label{fig:baseline-accuracy}
    \rmspace
\end{figure*}

Within each model family, larger models generally have higher accuracy but also higher energy per page (Figure~\ref{fig:baseline-accuracy}). Which architecture leads depends on the dataset: among the zero-shot models, Qwen3-VL and NuExtract are ahead on the layout-rich VRDU forms, while small text-only models with OCR are ahead on the text-heavy Kleister-NDA.

Arctic-TILT reports 92\% exact match on Kleister-NDA, but that figure is in-domain (Section~\ref{sec:datasets}) and not comparable to the zero-shot rows; we report it for reference and exclude it from best-configuration and Pareto claims on that dataset. Its VRDU behaviour is the informative comparison: 65\% exact match, below both Qwen3-VL-8B and NuExtract.

\subsection{RQ2 Input Representation}
Table~\ref{tab:energy_consumption_integrated} shows that whether a document is born-digital or scanned dictates the energy profile of the PDF-to-Markdown conversion. For born-digital documents (Kleister-NDA), Docling is the most efficient choice because it parses embedded text and bypasses visual OCR, which Tesseract ignores at a higher computational cost. For scanned documents (VRDU), Docling is forced to perform visual OCR, sharply increasing its energy footprint and making Tesseract the more efficient option. In both cases DeepSeek-OCR 2 consumes by far the most energy: $17\times$ Tesseract per page on Kleister-NDA and $18\times$ on VRDU.

These comparisons rest on whole-machine attribution, which does not treat every pipeline alike. Tesseract is measured without the GPU term (Section~\ref{sec:parsing}), so it is not charged for the idle L4; if anything this favors it. The GPU-accelerated parsers and all model inference include the GPU. CPU and RAM power are modeled rather than metered (RAM at a flat 20\,W), so the absolute energy gap between parser-heavy and vision-language pipelines carries some estimation error. The effects we draw conclusions from --- the ordering of the parsers within each dataset and the order-of-magnitude cost of neural OCR --- are far larger than that error.
\begin{table}[htb]
    \centering
    \footnotesize
    \caption{Total and per-page energy consumption (Wh) for PDF to Markdown conversion, over 500 documents per dataset. Per-page figures use each tool's own page count, which differs because Tesseract rasterizes the PDFs (3016 pages on Kleister-NDA, 1025 on VRDU) while Docling and DeepSeek-OCR~2 follow the PDF page structure (2934 and 915).}
    \label{tab:energy_consumption_integrated}
    \begin{tabular}{@{} l c c c c @{}}
        \toprule
        & \multicolumn{2}{c}{\textbf{Kleister-NDA}} & \multicolumn{2}{c}{\textbf{VRDU}} \\
        \cmidrule(lr){2-3} \cmidrule(lr){4-5}
        \textbf{Tool} & \textbf{Total} & \textbf{/Pg.} & \textbf{Total} & \textbf{/Pg.} \\
        \midrule
        Tesseract       & 12.91  & 0.0043 & 4.87  & 0.0047 \\
        Docling         & 4.38   & 0.0015 & 12.95 & 0.0141 \\
        DeepSeek-OCR 2  & 220.49 & 0.0751 & 76.41 & 0.0835 \\
        \bottomrule
    \end{tabular}
\end{table}

\begin{figure}[tb]
    \centering
    \includegraphics[width=\columnwidth]{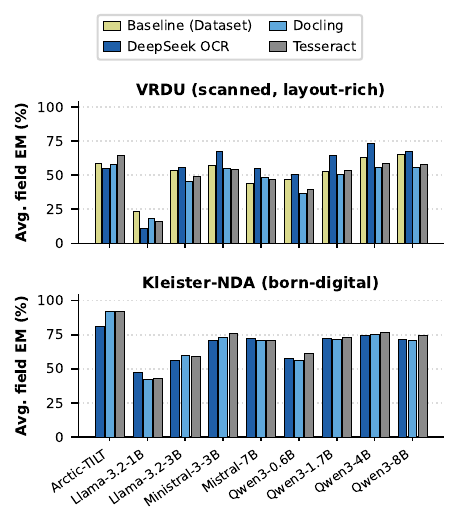}
    \caption{Average field exact match for each model under the four input representations (dataset-provided OCR, DeepSeek-OCR~2, Docling, Tesseract), at batch size 10 without quantization. Kleister-NDA ships no baseline OCR. Arctic-TILT's Kleister-NDA bars are in-domain (fine-tuned on that corpus). Underlying values are in Tables~\ref{tab:vrdu_plot_results} and~\ref{tab:kleister_nda_plot_results}.}
    \label{fig:input-modality-accuracy}
    \rmspace
\end{figure}

Extraction quality follows the same split (Figure~\ref{fig:input-modality-accuracy}). On the scanned, layout-rich VRDU forms, neural OCR (DeepSeek-OCR 2) is the most accurate representation for every general-purpose LLM except one, beating the better of Tesseract and Docling by 7--15 exact-match points. Two configurations run counter. Llama-3.2-1B, the smallest model, degrades sharply with DeepSeek-OCR 2 (16.5 EM with Tesseract to 11.2), suggesting its limited capacity is more sensitive to the longer, denser neural OCR output than to transcription quality. Arctic-TILT also performs worse, likely from spatial fidelity lost when approximating word-level bounding boxes from paragraph-level outputs (Appendix~\ref{sec:preprocessing}).
In contrast, for digital, text-heavy documents (Kleister-NDA), extraction accuracy varies by at most 6 points across the three parsing strategies for the general-purpose LLMs, with no parser consistently ahead. The exception is again Arctic-TILT, which spans 81.5--92.4 EM because it consumes bounding boxes rather than plain text and is therefore directly exposed to the layout fidelity of the parser. For text-only models on low-layout documents, then, a cheap parser or classical OCR suffices.

\subsection{RQ3 Inference Optimization}
\label{sec:rq3}
\begin{figure*}[ht]
    \centering
    \includegraphics[width=1.0\linewidth]{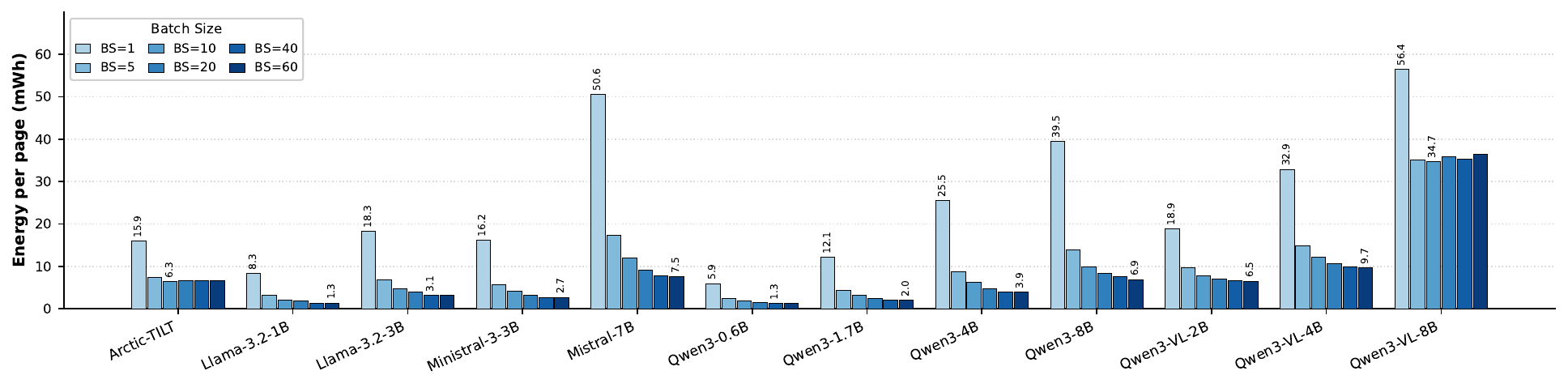}
    \caption{Inference energy per page on VRDU across models served with vLLM, unquantized, at batch sizes (BS) from 1 to 60; text-only models consume Tesseract output. Parsing energy is excluded. Data labels above the bars give the minimum and maximum energy per page observed for each model across the evaluated batch sizes.}
    \label{fig:energy-consumption-models-bs}
    \rmspace
\end{figure*}

Batching is the single most effective energy lever we measured (Figure~\ref{fig:energy-consumption-models-bs}). Moving from single-request serving to the best batch size reduces energy per page by 38--85\%, in this single-run sweep monotonically to batch size 60 for ten of the eleven models. Returns diminish steeply: most of the benefit is realized by batch size 10, and for Qwen3-8B energy drops from 39.5 to 9.8\,mWh/pg by batch size 10 but only to 6.9\,mWh/pg at batch size 60, so the last sixfold increase buys 8 of the 83 percentage points of savings.

The exception is the largest vision-language model. Qwen3-VL-8B reaches its minimum of 34.7\,mWh/pg already at batch size 10 and then flattens, holding between 35.3 and 36.4\,mWh/pg through batch size 60, because it sits close to the VRAM budget of the L4. Qwen3-VL-4B levels off in the same way once measured repeatedly: its single-run sweep keeps falling to batch size 60, but over three runs (below) its minimum is at batch size 20 (11.2\,mWh/pg), with a slight rise to 11.7 at batch size 60. The practical saturation point is therefore set by a model's memory headroom, not by a batch size that transfers across models.

\begin{figure}[ht]
    \centering
    \includegraphics[width=1.0\linewidth]{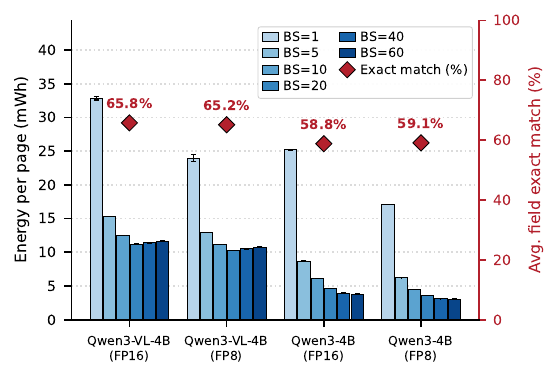}
    \caption{Energy per page and exact match for Qwen3-4B and Qwen3-VL-4B on VRDU, FP16 versus FP8, across batch sizes. Bars show the mean over three runs per configuration and error bars the standard deviation, which is mostly thinner than the bar outline. Exact match is averaged over runs and batch sizes, within which it varies by less than a point. It is Bench360's own field exact match rather than the recomputed metric of Tables~\ref{tab:vrdu_plot_results} and~\ref{tab:kleister_nda_plot_results}, and in these runs both models used the vision arm's sampling settings (Appendix~\ref{app:runconfig}), so absolute values differ slightly from the tables while the FP16-versus-FP8 difference is unaffected. The FP8 energy advantage is concentrated at batch size 1 and shrinks roughly tenfold by each model's best batch size.}
    \label{fig:accuracy-and-energy-quantization}
        \rmspace
\end{figure}

Figure~\ref{fig:accuracy-and-energy-quantization} compares unquantized (FP16) and FP8-quantized variants of Qwen3-4B and Qwen3-VL-4B across batch sizes on VRDU, each configuration measured three times; we report mean$\pm$standard deviation. At batch size~1, FP8 delivers substantial savings: 32.9$\pm$0.3 to 24.0$\pm$0.5\,mWh/pg for Qwen3-VL-4B ($-27\%$) and 25.2$\pm$0.1 to 17.1$\pm$0.1\,mWh/pg for Qwen3-4B ($-32\%$). Batching removes most of that advantage. At each model's best batch size the two precisions are 11.22$\pm$0.01 (FP16) against 10.26$\pm$0.02\,mWh/pg (FP8) for Qwen3-VL-4B, and 3.81$\pm$0.04 against 3.08$\pm$0.01\,mWh/pg for Qwen3-4B. FP8 therefore still saves 0.96 and 0.73\,mWh/pg ($-9\%$ and $-19\%$), a difference that is consistent across runs but roughly ten times smaller in absolute terms than at batch size~1. Accuracy is essentially unaffected: FP8 costs the vision model 0.6 exact-match points and gains the text model 0.3.

Batching and FP8 therefore act largely as substitutes rather than complements in this regime: both raise arithmetic intensity, and once batching has done so, low-precision arithmetic has much less idle capacity left to recover, consistent with the dequantization overheads reported by \citet{delavande2026understanding}. In a batched deployment FP8 is still worth a further 9--19\% of a much smaller energy budget, but its stronger argument is memory headroom.

\subsection{Pareto Frontier}
\label{sec:pareto}
\begin{figure*}[ht]
    \centering
    \includegraphics[width=\linewidth]{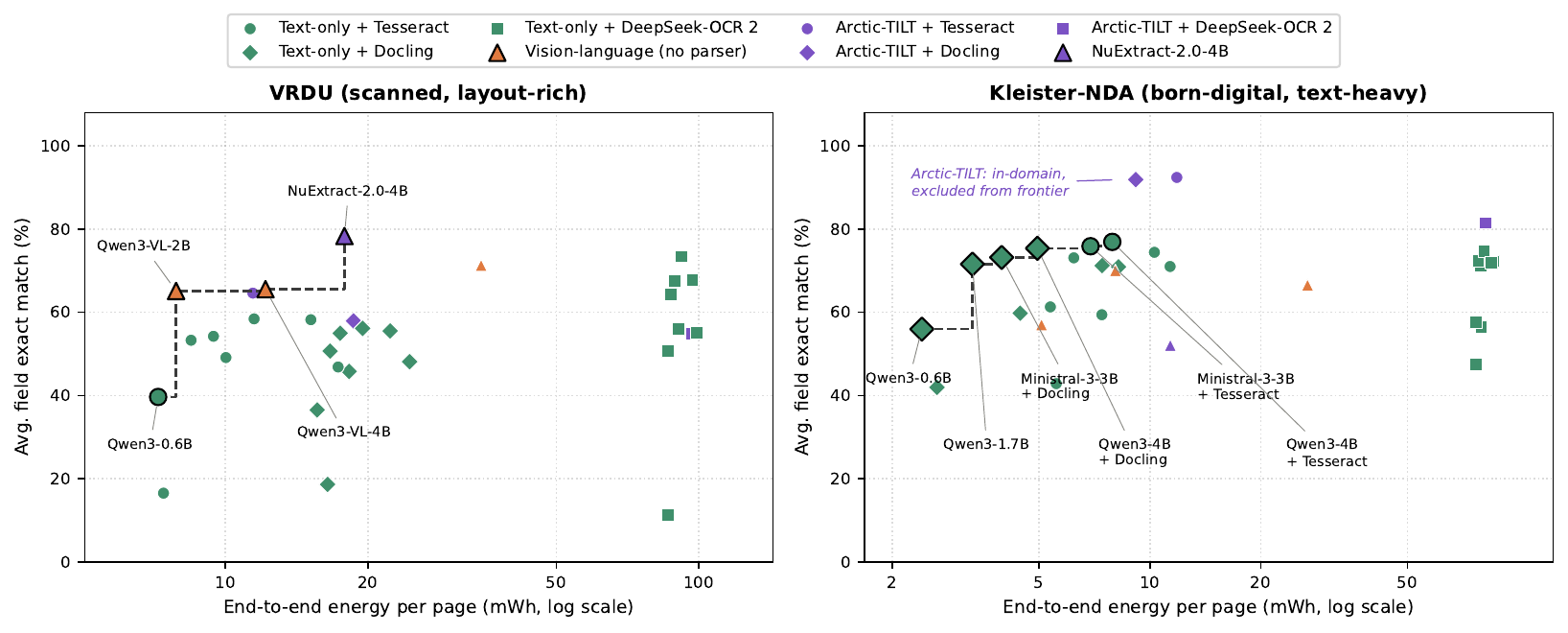}
    \caption{End-to-end accuracy--energy trade-off on VRDU (left) and Kleister-NDA (right). Energy per page includes both parsing and inference. Pareto-optimal configurations are enlarged, outlined in black, labelled, and joined by the dashed step line; all other measured configurations are shown at the same strength. The frontier is held by vision-language models on the layout-rich dataset and by text-only models with a cheap parser on the text-heavy one; no DeepSeek-OCR~2 configuration reaches either frontier. Frontiers cover zero-shot configurations only: Arctic-TILT is fine-tuned on Kleister-NDA and appears on that panel solely as an in-domain reference point, excluded from the frontier and from every zero-shot comparison.}
    \label{fig:pareto-combined}
    \rmspace
\end{figure*}

Figure~\ref{fig:pareto-combined} summarizes the end-to-end accuracy–energy trade-offs, making the impact of parsing energy clear. Text-only models paired with DeepSeek-OCR 2 are the most accurate text-only configurations on VRDU, but their parsing cost is an order of magnitude larger than the inference it feeds, and no DeepSeek-OCR 2 configuration reaches the frontier on either dataset. Every frontier and every comparison we draw on Kleister-NDA is strictly zero-shot: Arctic-TILT, fine-tuned on that corpus, is plotted on its panel only as an in-domain reference point and enters no frontier or zero-shot comparison (Section~\ref{sec:rq1}).

The composition of the frontier inverts between the two datasets. On VRDU it consists almost entirely of models that read page images directly: Qwen3-VL-2B (65.1\% EM at 7.9\,mWh/pg), Qwen3-VL-4B (65.6\% at 12.2\,mWh/pg) and NuExtract-2.0-4B (78.3\% at 17.8\,mWh/pg), with only the cheapest text-only configuration surviving at the low-accuracy end. On Kleister-NDA no vision-language model reaches the frontier at all; it is held end to end by text-only models paired with a cheap parser, from Qwen3-0.6B + Docling (56.0\% at 2.4\,mWh/pg) up to Qwen3-4B + Tesseract (76.9\% at 7.9\,mWh/pg), with Docling configurations occupying the low-energy half. The comparison is direct: on Kleister-NDA, Qwen3-4B + Docling is both more accurate and cheaper than Qwen3-VL-4B (75.4\% at 5.0\,mWh/pg against 70.1\% at 8.1\,mWh/pg), while on VRDU that ordering reverses (56.1\% at 19.5 against 65.6\% at 12.2).

Normalizing by page also makes the two corpora directly comparable, which per-document energy did not allow. A VRDU page is the more expensive of the two to process end to end: 11.5\,mWh/pg against 7.9 for Qwen3-4B with Tesseract, and 12.2 against 8.1 for Qwen3-VL-4B. The higher per-document cost of the contracts is therefore a length effect --- they run to three times as many pages --- rather than evidence that a contract page is harder to process than a scanned form page.

The same numbers bear on the length confound (Section~\ref{sec:datasets}). If the vision-language models lost on Kleister-NDA because long visual contexts scale poorly, their per-page energy would rise on the longer corpus; it falls instead (Qwen3-VL-4B: 8.1\,mWh/pg on Kleister-NDA against 12.2 on VRDU). The energy side of the inversion is therefore not a long-context effect. We cannot rule out a length effect on accuracy, and a length-matched subset would be needed to isolate it; our run reports store aggregate rather than per-document scores, so we leave this to future work.

Accuracy is also expensive at the margin. On VRDU, moving from Qwen3-VL-2B to Qwen3-VL-8B buys 6.2 exact-match points for $4.4\times$ the energy per page. Tables~\ref{tab:vrdu_plot_results} and~\ref{tab:kleister_nda_plot_results} in Appendix~\ref{detailed-results} give the underlying numbers for every configuration. 

\section{Discussion}
Naive single-document inference is the most expensive mistake available in this design space: batching alone removes 38--85\% of the energy per page, most of it by batch size 10. FP8 recovers much of the same energy when serving one request at a time but less than one milliwatt-hour per page once batching is in place, so the two are largely substitutes: stacking FP8 on batching still helps, but by roughly a tenth as much.

Parsing affects accuracy unevenly across document types, but affects energy uniformly and by far more. Embedded-text parsers land within a few points of the alternatives on born-digital contracts at a fraction of the cost. On scanned forms neural OCR does buy real accuracy, yet at $18\times$ Tesseract's per-page energy it costs an order of magnitude more than the inference it feeds, reaches neither Pareto frontier, and delivers accuracy that a vision-language model reading the page directly supplies more cheaply. A parser is charged for every page of every document while the accuracy it enables is capped by the model downstream, which makes parsing the first place to look when energy matters.

Which architecture wins depends on the type of document rather than on model size or family; in our corpora layout and length co-vary, so we attribute the inversion to the document type as a whole rather than to layout alone. Vision-language models are best placed on registration forms, reading page images and skipping parsing entirely; on low-layout contracts this reverses, and small text-only models with a cheap parser are both more accurate and cheaper than every vision-language configuration we measured. Arctic-TILT does not settle the question --- it leads on Kleister-NDA only because it was fine-tuned on it --- but, read as an in-domain upper bound, it does size the prize for domain adaptation, at roughly 15 exact-match points over the best zero-shot pipeline at comparable energy. Parsing, model selection, and serving configuration must be tuned as one design, not in isolation.

\section{Conclusion}
We compared local document IE pipelines from an accuracy--energy perspective at two ends of the layout spectrum, and derive three guidelines for building accurate, energy-efficient, and privacy-compliant pipelines with small local models. First, \emph{match parsing to layout:} embedded-text parsers or classical OCR for born-digital, text-heavy documents; neural OCR only for scanned forms, and only where its accuracy gain justifies an order of magnitude more parsing energy. Second, \emph{align architecture with structure:} vision-language models for layout-rich documents, where reading page images removes the parsing stage entirely; small text-only models with an efficient parser for low-layout documents, where they were both cheaper and more accurate; and, where in-domain data exists, adaptation of a small model in preference to scaling a zero-shot one at equal energy. Last, \emph{batch before you quantize:} increase batch size until memory pressure appears, worth 38--85\% of energy per page at no cost in accuracy, and treat FP8 as a smaller second step: it saves about 30\% for deployments that cannot batch but less than 1\,mWh per page on top of batching, where its main value is memory headroom.
\clearpage

\section*{Limitations}
\subsection*{Profiling of Energy Consumption}
Relying on CodeCarbon to measure the energy consumption is a limitation, as the profiling was conducted entirely using software-based estimation rather than direct hardware power meters. Because it is a software tool, it does not capture the total energy consumption \cite{Fischer_2026}. Systemic overheads, such as the energy required for hardware cooling, are excluded. This leads to an underestimation of the actual power consumption by 20-30\% when using CodeCarbon \cite{Fischer_2026}. However, because this underestimation represents a consistent deviation across all measurements, the comparative energy efficiency results and overall conclusions of this study remain robust.

Two further caveats apply. Parsing and specialized-model pipelines are profiled with CodeCarbon while vLLM-served models are profiled with Bench360. Both read the same NVML counter at the same 10\,Hz, so the GPU term is directly comparable; what we have not cross-validated is the modeled CPU and RAM contribution, which each tool estimates independently. In addition, CPU and RAM contributions are modeled from utilization rather than metered. Disabling GPU tracking for Tesseract keeps the idle L4 out of its figures, but the absolute gap between CPU-bound parsing and GPU inference still inherits the error of that model (Section~\ref{sec:parsing}).

\subsection*{Single Measurements}
The FP16-versus-FP8 comparison, whose differences at the best batch size fall below one milliwatt-hour per page, was measured three times per configuration. The standard deviation across runs was at most 0.5\,mWh/pg, and below 0.05 at the best batch sizes, so the remaining FP8 saving is well outside run-to-run noise. The repeats also show why single runs call for caution: the original Qwen3-VL-4B FP16 runs at batch sizes 40 and 60 lay 13--17\% below the repeat means. Every other configuration was measured once, apart from two that were run twice: exact match agreed within 0.9 and 2.2 points, and energy within 0.7\% and 8.8\%. The effects we base our recommendations on (38--85\% from batching, an order of magnitude between parsers) are well outside any plausible noise band.

\subsection*{Evaluation Protocol}
The text-only and vision-language arms use each family's recommended sampling defaults rather than one shared configuration. The bound in Section~\ref{sec:inference} ($-1.6$ to $+3.2$ exact-match points) shows that this is too small to explain the modality inversion. Greedy decoding would remove the difference entirely, together with a source of run-to-run variance, and is what we recommend for future benchmarks of this kind.

Arctic-TILT's Kleister-NDA results are in-domain: the released checkpoint is fine-tuned on that corpus, and we did not restrict our evaluation set to documents outside its training split, so the reported 92\% exact match should be read as an upper bound obtained with target-domain supervision rather than as a comparable system result. We exclude it from best-configuration and Pareto claims on Kleister-NDA for this reason, and note that a clean in-domain versus zero-shot comparison would require either the exact training split of the released model or a model we fine-tune ourselves. We also deviate from the VRDU benchmark protocol by evaluating zero-shot instead of few-shot fine-tuned.

Finally, our headline exact-match metric does not penalize fields a model leaves unanswered; the value-level F1 in Appendix~\ref{detailed-results} does, and the two should be read together.

\subsection*{Normalization and Document Length}
We report energy per page rather than per document, because the two corpora differ threefold in length and per-document figures would not be comparable across them. Two caveats follow. First, the divisor is a dataset-level average (5.87 and 1.83 pages per document), not a per-document page count, which the measurement data does not record. Dividing by it is therefore a constant rescale within each dataset, so every within-dataset result we report --- the batching and quantization effects, the model rankings, and the composition of both Pareto frontiers --- is identical to what per-document normalization would give. Only comparisons between the two datasets change.

Second, and more substantially, layout complexity and document length co-vary between our two corpora (Section~\ref{sec:datasets}), so the modality inversion is a property of the kind of document rather than of layout alone. Per-page energy argues against a long-context energy effect (Section~\ref{sec:pareto}), but length may still affect accuracy, for instance through the page cap on vision inputs. Separating the two factors would require a length-matched subset or a long-form layout-rich corpus, neither of which the available benchmarks provide. Pages are also an imperfect denominator for text-only models, whose energy scales with token count rather than page count.

\subsection*{Hardware Configuration}
The experiments were conducted using a NVIDIA L4 GPU (24 GB VRAM), which constrains the model selection. The available VRAM restricts evaluation to models of about $\leq 8\mathrm{B}$ parameters without quantization.
Prior work has shown that energy efficiency can vary substantially across GPU architectures and cooling solutions \cite{11553390,argerich2026wattcountsenergyawarebenchmark}, and predictive models such as WattGPU~\cite{argerich2026wattgpupredictinginferencepower} offer a way to estimate the inference side of our configurations on other GPUs without re-profiling. Thus, our findings should be interpreted as representative of a resource-constrained, single-GPU environment rather than universal across all hardware platforms. We expect the relative findings to transfer best: that batching saves energy by raising utilization, and the ordering of the parsers, which follows from how much work each does per page. The FP8 result is the most hardware-specific. The L4's Ada architecture has native FP8 tensor cores, whereas on GPUs without them vLLM falls back to weight-only FP8, so both the batch-size-1 saving and how much of it survives batching may differ on other platforms.

\subsection*{Quantization}
In this study, we focus on online FP8 quantization as implemented in vLLM, which applies quantization at inference time without modifying the underlying checkpoints. While this approach already yields notable energy savings, prior work on activation-aware weight quantization (AWQ) \cite{lin2023} suggests that carefully pre-quantized models can further reduce memory footprint and compute while preserving accuracy. Integrating such pre-quantized variants into our benchmarking framework is a promising direction for achieving even lower energy per request, especially for larger models that currently exceed the VRAM budget on our single L4 GPU.

\subsection*{Prompting Techniques}
The evaluations conducted in this study were limited to single-shot prompting. Advanced methods, such as chain-of-thought reasoning or other prompting strategies, were not evaluated. Consequently, the reported accuracy metrics should be interpreted as a lower bound of the models' capabilities.

\subsection*{Scope of the Energy Accounting}
Model loading is excluded from all reported figures: energy is measured over the generation phase only. Loading a model costs 142--259 seconds depending on size, which is substantial in absolute terms but is amortized away in the high-volume batch setting this paper targets. Deployments that load a model per request would need to account for it separately.

\subsection*{Coverage}
We study two English-language corpora representing two points on the layout spectrum. Invoices, receipts, multilingual forms such as the densely annotated regulatory documents of \citet{colakoglu2026agenticieadaptiveagentinformation}, and handwritten documents --- where the modality dependence we report would be stressed hardest --- are outside the scope of this evaluation. Adding such corpora, and in particular a long-form layout-rich one, is the most direct test of how far our guidelines generalize.

\section*{Ethical Considerations}
We are not aware of any ethical concerns or violations of the ACL Code of Ethics arising from this work. By empirically characterizing accuracy–energy trade-offs for local LLM inference, our study aims to support more responsible and accessible deployment of on-premise models.

\noindent\textbf{Use of Large Language Models (LLMs).} We used LLMs to assist with the writing of this manuscript. In particular, we drafted sections ourselves and then employed cloud-based LLM services (e.g., Gemini, ChatGPT) to refine wording and improve clarity, with all technical content and experimental claims authored and verified by the authors.

\clearpage

\bibliography{anthology,custom}

\begin{thebibliography}{39}
\providecommand{\natexlab}[1]{#1}

\bibitem[{Argerich et~al.(2026{\natexlab{a}})Argerich, Fürst, and
  Patiño-Martínez}]{argerich2026wattcountsenergyawarebenchmark}
Mauricio~Fadel Argerich, Jonathan Fürst, and Marta Patiño-Martínez.
  2026{\natexlab{a}}.
\newblock \href {https://arxiv.org/abs/2604.09048} {Watt counts: Energy-aware
  benchmark for sustainable {LLM} inference on heterogeneous {GPU}
  architectures}.
\newblock \emph{Preprint}, arXiv:2604.09048.

\bibitem[{Argerich et~al.(2026{\natexlab{b}})Argerich, Fürst, and
  Patiño-Martínez}]{argerich2026wattgpupredictinginferencepower}
Mauricio~Fadel Argerich, Jonathan Fürst, and Marta Patiño-Martínez.
  2026{\natexlab{b}}.
\newblock \href {https://arxiv.org/abs/2607.02391} {{WattGPU}: Predicting
  inference power and latency on unseen {GPUs} and {LLMs}}.
\newblock \emph{Preprint}, arXiv:2607.02391.

\bibitem[{Argerich and Patiño-Martínez(2024)}]{argerich2024measuring}
Mauricio~Fadel Argerich and Marta Patiño-Martínez. 2024.
\newblock \href {https://doi.org/10.1109/ACCESS.2024.3409745} {Measuring and
  improving the energy efficiency of large language models inference}.
\newblock \emph{IEEE Access}, 12:80194--80207.

\bibitem[{Borchmann et~al.(2025)Borchmann, Pietruszka, Jaskowski, Jurkiewicz,
  Halama, J{\'o}ziak, Garncarek, Liskowski, Szyndler, Gretkowski, O{\l}tusek,
  Nowakowska, Zaw{\l}ocki, Duhr, Dyda, and Turski}]{borchmann2025arctictilt}
{\L}ukasz Borchmann, Micha{\l} Pietruszka, Wojciech Jaskowski, Dawid
  Jurkiewicz, Piotr Halama, Pawe{\l} J{\'o}ziak, {\L}ukasz Garncarek, Pawe{\l}
  Liskowski, Karolina Szyndler, Andrzej Gretkowski, Julita O{\l}tusek, Gabriela
  Nowakowska, Artur Zaw{\l}ocki, {\L}ukasz Duhr, Pawe{\l} Dyda, and Micha{\l}
  Turski. 2025.
\newblock Arctic-tilt: Business document understanding at sub-billion scale.
\newblock In \emph{Proceedings of the 63rd Annual Meeting of the Association
  for Computational Linguistics (Volume 6: Industry Track)}, pages 264--283,
  Vienna, Austria. Association for Computational Linguistics.

\bibitem[{Borchmann et~al.(2021)Borchmann, Powalski, Stanislawek, Jurkiewicz,
  Stokowiec, Turski, Biecek, and Rokita}]{borchmann2021due}
{\L}ukasz Borchmann, Rafa{\l} Powalski, Tomasz Stanislawek, Dawid Jurkiewicz,
  Wojciech Stokowiec, Micha{\l} Turski, Przemys{\l}aw Biecek, and Przemys{\l}aw
  Rokita. 2021.
\newblock Due: End-to-end document understanding benchmark.
\newblock In \emph{NeurIPS 2021 Datasets and Benchmarks Track}.

\bibitem[{Caravaca(2026)}]{caravaca2026}
Francisco Caravaca. 2026.
\newblock \href {https://doi.org/10.1145/3788882.3788890} {Measuring energy
  consumption of llms inferences}.
\newblock \emph{SIGMETRICS Perform. Eval. Rev.}, 53(3):18–19.

\bibitem[{Colakoglu et~al.(2025)Colakoglu, Solmaz, and
  F{\"u}rst}]{colakoglu2025problem}
Gaye Colakoglu, G{\"u}rkan Solmaz, and Jonathan F{\"u}rst. 2025.
\newblock Problem solved? information extraction design space for layout-rich
  documents using llms.
\newblock In \emph{EMNLP (Findings)}, pages 17908--17927.

\bibitem[{Colakoglu et~al.(2026)Colakoglu, Solmaz, and
  Fürst}]{colakoglu2026agenticieadaptiveagentinformation}
Gaye Colakoglu, Gürkan Solmaz, and Jonathan Fürst. 2026.
\newblock \href {https://arxiv.org/abs/2509.11773} {{AgenticIE}: An adaptive
  agent for information extraction from complex regulatory documents}.
\newblock \emph{Preprint}, arXiv:2509.11773.

\bibitem[{Courty et~al.(2024)Courty, Schmidt, Luccioni, Goyal-Kamal,
  MarionCoutarel, Feld, Lecourt, LiamConnell, Saboni, Inimaz, supatomic,
  Léval, Blanche, Cruveiller, ouminasara, Zhao, Joshi, Bogroff, de~Lavoreille,
  Laskaris, Abati, Blank, Wang, Catovic, Alencon, Stęchły, Bauer, de~Araújo,
  JPW, and MinervaBooks}]{benoit_courty_2024_11171501}
Benoit Courty, Victor Schmidt, Sasha Luccioni, Goyal-Kamal, MarionCoutarel,
  Boris Feld, Jérémy Lecourt, LiamConnell, Amine Saboni, Inimaz, supatomic,
  Mathilde Léval, Luis Blanche, Alexis Cruveiller, ouminasara, Franklin Zhao,
  Aditya Joshi, Alexis Bogroff, Hugues de~Lavoreille, and 11 others. 2024.
\newblock \href {https://doi.org/10.5281/zenodo.11171501} {mlco2/codecarbon:
  v2.4.1}.

\bibitem[{{DeepSeek-AI}(2026)}]{deepseekocr2}
{DeepSeek-AI}. 2026.
\newblock \href {https://arxiv.org/abs/2601.20552} {{DeepSeek-OCR} 2: Visual
  causal flow}.
\newblock \emph{Preprint}, arXiv:2601.20552.

\bibitem[{Delavande et~al.(2026)Delavande, Pierrard, and
  Luccioni}]{delavande2026understanding}
Julien Delavande, Regis Pierrard, and Sasha Luccioni. 2026.
\newblock \href {https://arxiv.org/abs/2601.22362} {Understanding efficiency:
  Quantization, batching, and serving strategies in llm energy use}.
\newblock \emph{Preprint}, arXiv:2601.22362.

\bibitem[{Denk and Reisswig(2019)}]{denk2019bertgrid}
Timo~I. Denk and Christian Reisswig. 2019.
\newblock \href {https://arxiv.org/abs/1909.04948} {Bertgrid: Contextualized
  embedding for 2d document representation and understanding}.
\newblock \emph{Preprint}, arXiv:1909.04948.

\bibitem[{Ding et~al.(2025)Ding, Han, Lee, and Hovy}]{ding2024survey}
Yihao Ding, Soyeon~Caren Han, Jean Lee, and Eduard Hovy. 2025.
\newblock \href {https://arxiv.org/abs/2408.01287} {Deep learning based
  visually rich document content understanding: A survey}.
\newblock \emph{Preprint}, arXiv:2408.01287.

\bibitem[{Fernandez et~al.(2025)Fernandez, Na, Tiwari, Bisk, Luccioni, and
  Strubell}]{fernandez2025energy}
Jared Fernandez, Clara Na, Vashisth Tiwari, Yonatan Bisk, Sasha Luccioni, and
  Emma Strubell. 2025.
\newblock \href {https://doi.org/10.18653/v1/2025.acl-long.1563} {Energy
  considerations of large language model inference and efficiency
  optimizations}.
\newblock In \emph{Proceedings of the 63rd Annual Meeting of the Association
  for Computational Linguistics (Volume 1: Long Papers)}, pages 32556--32569,
  Vienna, Austria. Association for Computational Linguistics.

\bibitem[{Fischer(2026)}]{Fischer_2026}
Raphael Fischer. 2026.
\newblock \href {https://doi.org/10.1515/itit-2025-0031} {Ground-truthing ai
  energy consumption: validating codecarbon against external measurements}.
\newblock \emph{it - Information Technology}.

\bibitem[{Grattafiori et~al.(2024)Grattafiori, Dubey et~al.}]{llama3}
Aaron Grattafiori, Abhimanyu Dubey, and 1 others. 2024.
\newblock \href {https://arxiv.org/abs/2407.21783} {The {Llama} 3 herd of
  models}.
\newblock \emph{Preprint}, arXiv:2407.21783.

\bibitem[{Jiang et~al.(2023)Jiang, Sablayrolles, Mensch et~al.}]{mistral7b}
Albert~Q. Jiang, Alexandre Sablayrolles, Arthur Mensch, and 1 others. 2023.
\newblock \href {https://arxiv.org/abs/2310.06825} {{Mistral} 7{B}}.
\newblock \emph{Preprint}, arXiv:2310.06825.

\bibitem[{Katti et~al.(2018)Katti, Reisswig, Guder, Brarda, Bickel, Höhne, and
  Faddoul}]{katti2018chargrid}
Anoop~Raveendra Katti, Christian Reisswig, Cordula Guder, Sebastian Brarda,
  Steffen Bickel, Johannes Höhne, and Jean~Baptiste Faddoul. 2018.
\newblock \href {https://arxiv.org/abs/1809.08799} {Chargrid: Towards
  understanding 2d documents}.
\newblock \emph{Preprint}, arXiv:1809.08799.

\bibitem[{Kim et~al.(2022)Kim, Hong, Yim, Nam, Park, Yim, Hwang, Yun, Han, and
  Park}]{kim2022donut}
Geewook Kim, Teakgyu Hong, Moonbin Yim, JeongYeon Nam, Jinyoung Park, Jinyeong
  Yim, Wonseok Hwang, Sangdoo Yun, Dongyoon Han, and Seunghyun Park. 2022.
\newblock {OCR}-free document understanding transformer.
\newblock In \emph{European Conference on Computer Vision (ECCV)}.

\bibitem[{Kwon et~al.(2023)Kwon, Li, Zhuang, Sheng, Zheng, Yu, Gonzalez, Zhang,
  and Stoica}]{kwon2023vllm}
Woosuk Kwon, Zhuohan Li, Siyuan Zhuang, Ying Sheng, Lianmin Zheng, Cody~Hao Yu,
  Joseph~E. Gonzalez, Hao Zhang, and Ion Stoica. 2023.
\newblock Efficient memory management for large language model serving with
  {PagedAttention}.
\newblock In \emph{Proceedings of the 29th Symposium on Operating Systems
  Principles (SOSP)}.

\bibitem[{Latif et~al.(2025)Latif, Shafique, Ullah, Newkirk, Yu, and
  Munir}]{11553390}
Imran Latif, Muhammad~Ali Shafique, Hayat Ullah, Alex~C. Newkirk, Xi~Yu, and
  Arslan Munir. 2025.
\newblock \href {https://arxiv.org/abs/2507.16781} {Cooling matters:
  Benchmarking large language models and vision-language models on
  liquid-cooled versus air-cooled h100 gpu systems}.
\newblock \emph{Preprint}, arXiv:2507.16781.

\bibitem[{Lee et~al.(2023)Lee, Joshi, Turc, Hu, Liu, Eisenschlos, Khandelwal,
  Shaw, Chang, and Toutanova}]{lee2023pix2struct}
Kenton Lee, Mandar Joshi, Iulia Turc, Hexiang Hu, Fangyu Liu, Julian
  Eisenschlos, Urvashi Khandelwal, Peter Shaw, Ming-Wei Chang, and Kristina
  Toutanova. 2023.
\newblock {Pix2Struct}: Screenshot parsing as pretraining for visual language
  understanding.
\newblock In \emph{International Conference on Machine Learning (ICML)}.

\bibitem[{Lin et~al.(2026)Lin, Tang, Tang, Yang, Chen, Wang, Xiao, Dang, Gan,
  and Han}]{lin2023}
Ji~Lin, Jiaming Tang, Haotian Tang, Shang Yang, Wei-Ming Chen, Wei-Chen Wang,
  Guangxuan Xiao, Xingyu Dang, Chuang Gan, and Song Han. 2026.
\newblock \href {https://arxiv.org/abs/2306.00978} {Awq: Activation-aware
  weight quantization for llm compression and acceleration}.
\newblock \emph{Preprint}, arXiv:2306.00978.

\bibitem[{Livathinos et~al.(2025)Livathinos, Auer, Lysak, Nassar, Dolfi,
  Vagenas, Ramis, Omenetti, Dinkla, Kim, Gupta, de~Lima, Weber, Morin, Meijer,
  Kuropiatnyk, and Staar}]{Livathinosetal2025}
Nikolaos Livathinos, Christoph Auer, Maksym Lysak, Ahmed Nassar, Michele Dolfi,
  Panos Vagenas, Cesar~Berrospi Ramis, Matteo Omenetti, Kasper Dinkla, Yusik
  Kim, Shubham Gupta, Rafael~Teixeira de~Lima, Valery Weber, Lucas Morin,
  Ingmar Meijer, Viktor Kuropiatnyk, and Peter W.~J. Staar. 2025.
\newblock \href {https://arxiv.org/abs/2501.17887} {Docling: An efficient
  open-source toolkit for ai-driven document conversion}.
\newblock \emph{Preprint}, arXiv:2501.17887.

\bibitem[{Luccioni et~al.(2024)Luccioni, Jernite, and
  Strubell}]{luccioni2024power}
Alexandra~Sasha Luccioni, Yacine Jernite, and Emma Strubell. 2024.
\newblock Power hungry processing: {Watts} driving the cost of {AI} deployment?
\newblock In \emph{Proceedings of the 2024 ACM Conference on Fairness,
  Accountability, and Transparency (FAccT)}.

\bibitem[{Luo et~al.(2024)Luo, Shen, Zhu, Zheng, Yu, and
  Yao}]{luo2024layoutllm}
Chuwei Luo, Yufan Shen, Zhaoqing Zhu, Qi~Zheng, Zhi Yu, and Cong Yao. 2024.
\newblock \href {https://arxiv.org/abs/2404.05225} {Layoutllm: Layout
  instruction tuning with large language models for document understanding}.
\newblock \emph{Preprint}, arXiv:2404.05225.

\bibitem[{{Mistral AI}(2025)}]{ministral3}
{Mistral AI}. 2025.
\newblock \href {https://huggingface.co/mistralai/Ministral-3-3B-Instruct-2512}
  {{Ministral-3-3B-Instruct-2512}}.
\newblock Model weights, Hugging Face.

\bibitem[{NuMind(2026)}]{nuextract2}
NuMind. 2026.
\newblock \href {https://nuextract.ai/} {Nuextract 2.0}.

\bibitem[{Poddar et~al.(2025)Poddar, Koley, Misra, Podder, Ganguly, and
  Ghosh}]{poddar2025benchmarking}
Soham Poddar, Paramita Koley, Janardan Misra, Sanjay Podder, Niloy Ganguly, and
  Saptarshi Ghosh. 2025.
\newblock \href {https://arxiv.org/abs/2502.05610} {Towards sustainable nlp:
  Insights from benchmarking inference energy in large language models}.
\newblock \emph{Preprint}, arXiv:2502.05610.

\bibitem[{Pronk and Zhao(2025)}]{pronk2025benchmarkingenergyefficiencylarge}
K.~Pronk and Q.~Zhao. 2025.
\newblock \href {https://arxiv.org/abs/2509.08867} {Benchmarking energy
  efficiency of large language models using vllm}.
\newblock \emph{Preprint}, arXiv:2509.08867.

\bibitem[{{Qwen Team}(2025{\natexlab{a}})}]{qwen3}
{Qwen Team}. 2025{\natexlab{a}}.
\newblock \href {https://arxiv.org/abs/2505.09388} {{Qwen3} technical report}.
\newblock \emph{Preprint}, arXiv:2505.09388.

\bibitem[{{Qwen Team}(2025{\natexlab{b}})}]{qwen3vl}
{Qwen Team}. 2025{\natexlab{b}}.
\newblock \href {https://huggingface.co/Qwen} {{Qwen3-VL}}.
\newblock Model weights, Hugging Face.

\bibitem[{Smith(2007)}]{smith2007tesseract}
Ray Smith. 2007.
\newblock An overview of the {Tesseract} {OCR} engine.
\newblock In \emph{Ninth International Conference on Document Analysis and
  Recognition (ICDAR)}, pages 629--633.

\bibitem[{Stanisławek et~al.(2021)Stanisławek, Graliński, Wróblewska,
  Lipiński, Kaliska, Rosalska, Topolski, and Biecek}]{kleister-nda}
Tomasz Stanisławek, Filip Graliński, Anna Wróblewska, Dawid Lipiński,
  Agnieszka Kaliska, Paulina Rosalska, Bartosz Topolski, and Przemysław
  Biecek. 2021.
\newblock \href {https://doi.org/10.1007/978-3-030-86549-8_36} {\emph{Kleister:
  Key Information Extraction Datasets Involving Long Documents with Complex
  Layouts}}, page 564–579.
\newblock Springer International Publishing.

\bibitem[{Stuhlmann et~al.(2026)Stuhlmann, Argerich, and
  Fürst}]{stuhlmann2026bench360benchmarkinglocalllm}
Linus Stuhlmann, Mauricio~Fadel Argerich, and Jonathan Fürst. 2026.
\newblock \href {https://arxiv.org/abs/2511.16682} {Bench360: Benchmarking
  local llm inference from 360 degrees}.
\newblock \emph{Preprint}, arXiv:2511.16682.

\bibitem[{Wang et~al.(2023)Wang, Zhou, Wei, Lee, and Tata}]{vrdu-dataset}
Zilong Wang, Yichao Zhou, Wei Wei, Chen-Yu Lee, and Sandeep Tata. 2023.
\newblock \href {https://doi.org/10.1145/3580305.3599929} {Vrdu: A benchmark
  for visually-rich document understanding}.
\newblock In \emph{Proceedings of the 29th ACM SIGKDD Conference on Knowledge
  Discovery and Data Mining}, KDD ’23, page 5184–5193. ACM.

\bibitem[{Watanabe et~al.(1995)Watanabe, Luo, and Sugie}]{watanabe1995layout}
Toyohide Watanabe, Qin Luo, and Noboru Sugie. 1995.
\newblock Layout recognition of multi-kinds of table-form documents.
\newblock \emph{IEEE Transactions on Pattern Analysis and Machine
  Intelligence}, 17(4):432--445.

\bibitem[{Xu et~al.(2024)Xu, Chen, Peng, Zhang, Xu, Zhao, Wu, Zheng, Wang, and
  Chen}]{xu2024llm-ie}
Derong Xu, Wei Chen, Wenjun Peng, Chao Zhang, Tong Xu, Xiangyu Zhao, Xian Wu,
  Yefeng Zheng, Yang Wang, and Enhong Chen. 2024.
\newblock \href {https://arxiv.org/abs/2312.17617} {Large language models for
  generative information extraction: A survey}.
\newblock \emph{Preprint}, arXiv:2312.17617.

\bibitem[{Xu et~al.(2020)Xu, Li, Cui, Huang, Wei, and Zhou}]{xu2020layoutlm}
Yiheng Xu, Minghao Li, Lei Cui, Shaohan Huang, Furu Wei, and Ming Zhou. 2020.
\newblock \href {https://doi.org/10.1145/3394486.3403172} {Layoutlm:
  Pre-training of text and layout for document image understanding}.
\newblock In \emph{Proceedings of the 26th ACM SIGKDD International Conference
  on Knowledge Discovery \& Data Mining}, KDD ’20, page 1192–1200. ACM.

\end{thebibliography}
\clearpage

\appendix
\section{Run Configuration}
\label{app:runconfig}

Decoding parameters follow each model family's recommended defaults and the token budget of the dataset:

\begin{center}
\footnotesize
\begin{tabular}{@{}llccc@{}}
\toprule
Arm & Dataset & temp. & top-$p$ & max tok. \\
\midrule
Text-only & Kleister-NDA & 0.6 & 0.9 & 128 \\
Text-only & VRDU & 0.6 & 0.9 & 256 \\
Vision & Kleister-NDA & 0.7 & 0.8 & 128 \\
Vision & VRDU & 0.7 & 0.8 & 128 \\
\bottomrule
\end{tabular}
\end{center}

\noindent Observed generation lengths averaged between 7.4 and 25.9 tokens across every run, so no configuration was budget-limited. For all Qwen3 and Qwen3-VL models, reasoning traces were disabled by passing \texttt{chat\_template\_kwargs=\allowbreak\{"enable\_\allowbreak thinking":\ false\}}; the Mistral and Ministral models do not expose such a mode. Vision-language runs pass at most the first 10 (Kleister-NDA) or 5 (VRDU) page images per document.

\noindent\textbf{NuExtract}
Configuration of vLLM for NuExtract 2.0 4B:
\begin{verbatim}
vllm/vllm-openai:latest \
  --model numind/NuExtract-2.0-4B \
  --trust-remote-code \
  --dtype bfloat16 \
  --gpu-memory-utilization 0.8 \
  --max-model-len 32768 \
  --chat-template-content-format openai
\end{verbatim}

\noindent\textbf{Text-only models without quantization (vLLM).}
For text-only runs without quantization, vLLM was configured as follows:
\begin{verbatim}
vllm/vllm-openai:latest \
    --model "$MODEL" \
    --trust-remote-code \
    --max-model-len 31872 \
    --gpu-memory-utilization 0.95 \
    --port "$PORT"
\end{verbatim}

\noindent\textbf{Text-only models with FP8 quantization (vLLM).}
For text-only runs with FP8 quantization, we used:
\begin{verbatim}
vllm/vllm-openai:latest \
    --model "$MODEL" \
    --trust-remote-code \
    --max-model-len 31872 \
    --gpu-memory-utilization 0.95 \
    --port "$PORT" \
    --quantization fp8
\end{verbatim}

\noindent\textbf{Vision models without quantization (vLLM).}
For multimodal (vision) models without quantization, vLLM was configured as:
\begin{verbatim}
vllm/vllm-openai:latest \
    --model "$MODEL" \
    --trust-remote-code \
    --max-model-len 28000 \
    --port "$PORT" \
    --gpu-memory-utilization 0.95 \
    --no-enable-prefix-caching \
    --limit-mm-per-prompt '{"image": 15}'
\end{verbatim}

\noindent\textbf{Vision models with FP8 quantization (vLLM).}
For multimodal (vision) models with FP8 quantization, we used:
\begin{verbatim}
vllm/vllm-openai:latest \
    --model "$MODEL" \
    --trust-remote-code \
    --max-model-len 28000 \
    --port "$PORT" \
    --gpu-memory-utilization 0.95 \
    --no-enable-prefix-caching \
    --quantization fp8 \
    --limit-mm-per-prompt '{"image": 15}'
\end{verbatim}

\noindent\textbf{Arctic TILT configuration.}
For Arctic TILT, with \verb|$BS| denoting the batch size, we used:

\begin{verbatim}
python examples/tilt_example.py \
    --model Snowflake/snowflake
    -arctic-tilt-v1.3 \
    --dataset "$DATASET" \
    --output-dir "$OUTPUT_DIR" \
    --gpu-memory-utilization 0.8 \
    --subset "" \
    --limit-documents 500 \
    --async \
    --max-num-seqs $BS \
    --enforce-eager
\end{verbatim}

\subsection{Bounding-box preprocessing for Arctic-TILT}
\label{sec:preprocessing}
Arctic-TILT \cite{borchmann2025arctictilt} expects documents in the Document Understanding (DU) schema \cite{borchmann2021due}, which requires word-level bounding boxes. DeepSeek-OCR 2 emits bounding boxes at paragraph level instead. We approximate word-level boxes by interpolating their coordinates proportionally from the position of each token within the text of the enclosing paragraph box.

\section{Appendix: Prompt Templates}
\label{app:prompt_templates}

This section details the prompt templates used for the different information extraction tasks.

\subsection{VRDU Text Information Extraction}
For text-based extraction on the VRDU dataset, the system message establishes strict JSON output rules, and the user prompt directly injects the OCR text.

\noindent\textbf{System Prompt:}
\begin{lstlisting}
You are an information extraction engine.
Your task is to read OCR text from a document and extract specific fields.
You must output ONLY one JSON object, with EXACTLY the requested keys.
Rules:
  - The JSON must be on a single line (no line breaks or indentation).
  - Each requested key MUST be present in the JSON.
  - If a field appears multiple times, use a JSON array of unique values in reading order.
  - If a field is not present in the OCR text, set its value to null.
  - Do NOT add any keys that were not requested.
  - Do NOT output any explanations, comments, or text outside the JSON object.
\end{lstlisting}

\noindent\textbf{User Prompt:}
\begin{lstlisting}
Extract the requested keys from the OCR
OCR:
{ocr_text}

Requested keys:
{fields_str}

Output JSON (single line, no extra text):
\end{lstlisting}

\subsection{VRDU Visual Information Extraction}
For Vision-Language Models processing VRDU document images, the system message is minimized. The user prompt precedes a sequence of Base64-encoded images.

\noindent\textbf{System Prompt:}
\begin{lstlisting}
You are an information extraction engine. Output ONLY JSON.
\end{lstlisting}

\noindent\textbf{User Prompt:}
\begin{lstlisting}
Extract these fields from the images: {fields_str}
Output JSON on one line.
\end{lstlisting}
\textit{Note: This text is followed by the Base64 image payloads.}

\subsection*{Dynamic Field Guidelines for Kleister-NDA}
In the Kleister-NDA prompt templates below, the \texttt{\{requested\_descriptions\}} placeholder is dynamically populated. To avoid confusing the model with irrelevant instructions, the pipeline only injects guidelines for fields that are actually present in the ground truth of the specific document being evaluated. 

The possible field guidelines that can be injected are:
\begin{itemize}
    \item \textbf{effective\_date}: Extract the effective date. Strictly format as YYYY-MM-DD.
    \item \textbf{jurisdiction}: The state or country whose laws govern the agreement (e.g., 'New York', 'Delaware').
    \item \textbf{party}: The exact names of the companies, organizations, or individuals entering into the agreement.
    \item \textbf{term}: The duration of the agreement. Format as a number followed by the unit (e.g., '3 years', '1 year', '6 months').
\end{itemize}

\subsection{Kleister-NDA Text Extraction}
For text-based extraction on the Kleister-NDA dataset, the system message, hardcoded field guidelines, and the user prompt containing the document text are concatenated into a single block.

\noindent\textbf{Combined Prompt:}
\begin{lstlisting}
You are an information extraction engine.
Your task is to extract specific fields from the provided Non-Disclosure Agreement (NDA) document text.
You must output ONLY one JSON object, with EXACTLY the requested keys.
Rules:
  - The JSON must be on a single line (no line breaks or indentation).
  - Each requested key MUST be present in the JSON.
  - If a field appears multiple times (e.g., multiple parties), use a JSON array of unique values.
  - Do NOT output any explanations, comments, or text outside the JSON object.

Field Guidelines:
{requested_descriptions}

--- DOCUMENT TEXT ---
{document_text}
--- END OF DOCUMENT ---

Requested keys:
{fields_str}

Output JSON:
\end{lstlisting}

\subsection{Kleister-NDA Visual Extraction}
For Vision-Language Models processing Kleister-NDA document images, the system message dynamically includes field guidelines only for the requested keys. The user prompt precedes the Base64-encoded images.

\noindent\textbf{System Prompt:}
\begin{lstlisting}
You are an information extraction engine.
Your task is to extract specific fields from the provided Non-Disclosure Agreement (NDA) document images.
You must output ONLY one JSON object, with EXACTLY the requested keys.
Rules:
  - The JSON must be on a single line (no line breaks or indentation).
  - Each requested key MUST be present in the JSON.
  - If a field appears multiple times (e.g., multiple parties), use a JSON array of unique values.
  - Do NOT output any explanations, comments, or text outside the JSON object.

Field Guidelines:
{requested_descriptions}
\end{lstlisting}

\noindent\textbf{User Prompt:}
\begin{lstlisting}
Extract these fields from the images: {fields_str}
Output JSON on one line.
\end{lstlisting}
\textit{Note: This text is followed by the Base64 image payloads.}

\section{Detailed Results}

\noindent\textbf{Macro F1 Score (F1):} Evaluates the overall extraction performance by calculating the harmonic mean of precision and recall. This is computed by assessing the value-level intersection of predicted and ground-truth sets for each document, which is then averaged across the entire dataset.

\noindent\textbf{Fuzzy Match Score (FM):} Accounts for minor OCR errors or generative variations by applying a token-sort ratio algorithm. This calculates a similarity score between 0 and 1 by alphabetically sorting tokens before comparison, ensuring robust matching even if the model outputs words in a slightly different order.

\label{detailed-results}
\begin{table*}[ht]
\caption{Detailed benchmarking results for the VRDU dataset (Pareto plot configurations), over 500 documents. OCR~(mWh/pg) is the parsing energy per page and E2E~(mWh/pg) additionally includes model inference; both are normalized by the 1.83 pages per document of this dataset. The OCR text shipped with the VRDU benchmark is used for accuracy comparisons in Figure~\ref{fig:input-modality-accuracy} but excluded here, since the energy of producing it is not attributable to our pipeline.}
\label{tab:vrdu_plot_results}
\begin{tabular}{llcccccc}
\toprule
& & & & & & \multicolumn{2}{c}{Energy (mWh/pg)} \\
\cmidrule(lr){7-8}
Model & BS & OCR & EM (\%) & FM (\%) & F1 (\%) & OCR & E2E \\
\midrule
Arctic-TILT & 10 & DeepSeek & 54.8 & 74.5 & 55.0 & 83.5 & 96.8 \\
Arctic-TILT & 10 & Docling & 57.9 & 74.0 & 58.0 & 14.1 & 18.7 \\
Arctic-TILT & 10 & Tesseract & 64.6 & 74.7 & 64.6 & 5.3 & 11.4 \\
Llama-3.2-1B & 10 & DeepSeek & 11.2 & 13.4 & 9.8 & 83.5 & 86.0 \\
Llama-3.2-1B & 10 & Docling & 18.6 & 31.2 & 17.6 & 14.1 & 16.4 \\
Llama-3.2-1B & 10 & Tesseract & 16.5 & 23.3 & 15.6 & 5.3 & 7.4 \\
Llama-3.2-3B & 10 & DeepSeek & 55.9 & 77.5 & 57.4 & 83.5 & 90.7 \\
Llama-3.2-3B & 10 & Docling & 45.8 & 71.6 & 45.4 & 14.1 & 18.3 \\
Llama-3.2-3B & 10 & Tesseract & 49.1 & 73.1 & 49.3 & 5.3 & 10.0 \\
Ministral-3-3B & 10 & DeepSeek & 67.6 & 80.3 & 64.8 & 83.5 & 88.9 \\
Ministral-3-3B & 10 & Docling & 55.0 & 69.6 & 51.2 & 14.1 & 17.5 \\
Ministral-3-3B & 10 & Tesseract & 54.2 & 71.1 & 52.7 & 5.3 & 9.4 \\
Mistral-7B & 10 & DeepSeek & 55.1 & 71.0 & 53.1 & 83.5 & 99.0 \\
Mistral-7B & 10 & Docling & 48.1 & 67.2 & 45.7 & 14.1 & 24.5 \\
Mistral-7B & 10 & Tesseract & 46.9 & 63.4 & 45.3 & 5.3 & 17.3 \\
NuExtract-2.0-4B & 1 & - & 78.3 & 81.5 & 74.0 & - & 17.8 \\
Qwen3-0.6B & 10 & DeepSeek & 50.7 & 70.2 & 50.7 & 83.5 & 86.0 \\
Qwen3-0.6B & 10 & Docling & 36.5 & 64.4 & 36.5 & 14.1 & 15.6 \\
Qwen3-0.6B & 10 & Tesseract & 39.6 & 63.9 & 39.7 & 5.3 & 7.2 \\
Qwen3-1.7B & 10 & DeepSeek & 64.4 & 77.3 & 61.3 & 83.5 & 87.4 \\
Qwen3-1.7B & 10 & Docling & 50.7 & 70.4 & 48.7 & 14.1 & 16.6 \\
Qwen3-1.7B & 10 & Tesseract & 53.3 & 74.8 & 52.9 & 5.3 & 8.5 \\
Qwen3-4B & 10 & DeepSeek & 73.4 & 84.9 & 70.9 & 83.5 & 91.8 \\
Qwen3-4B & 10 & Docling & 56.1 & 75.0 & 53.8 & 14.1 & 19.5 \\
Qwen3-4B & 10 & Tesseract & 58.4 & 78.7 & 57.3 & 5.3 & 11.5 \\
Qwen3-8B & 10 & DeepSeek & 67.8 & 82.7 & 65.2 & 83.5 & 97.0 \\
Qwen3-8B & 10 & Docling & 55.5 & 72.2 & 51.9 & 14.1 & 22.3 \\
Qwen3-8B & 10 & Tesseract & 58.2 & 76.9 & 56.0 & 5.3 & 15.2 \\
Qwen3-VL-2B & 10 & - & 65.1 & 82.7 & 65.2 & - & 7.9 \\
Qwen3-VL-4B & 10 & - & 65.6 & 83.6 & 65.6 & - & 12.2 \\
Qwen3-VL-8B & 10 & - & 71.3 & 86.2 & 71.4 & - & 34.7 \\
\bottomrule
\end{tabular}
\end{table*}

\begin{table*}[t]
\caption{Detailed benchmarking results for the Kleister-NDA dataset (Pareto plot configurations). Extraction is evaluated over 337 documents; parsing energy is amortized over the 500 documents of the parsing benchmark. Energy is per page, normalized by the 5.87 pages per document of this dataset. Arctic-TILT is fine-tuned on Kleister-NDA by its authors and is therefore evaluated in-domain, unlike every other model in the table; its rows are reported for reference and are excluded from best-configuration and Pareto comparisons.}
\label{tab:kleister_nda_plot_results}
\begin{tabular}{llcccccc}
\toprule
& & & & & & \multicolumn{2}{c}{Energy (mWh/pg)} \\
\cmidrule(lr){7-8}
Model & BS & OCR & EM (\%) & FM (\%) & F1 (\%) & OCR & E2E \\
\midrule
Arctic-TILT$^\dagger$ & 10 & DeepSeek & 81.5 & 91.2 & 84.8 & 75.1 & 81.3 \\
Arctic-TILT$^\dagger$ & 10 & Docling & 91.9 & 98.0 & 93.8 & 1.5 & 9.2 \\
Arctic-TILT$^\dagger$ & 10 & Tesseract & 92.4 & 98.1 & 93.8 & 4.4 & 11.8 \\
Llama-3.2-1B & 10 & DeepSeek & 47.6 & 70.7 & 58.4 & 75.1 & 76.7 \\
Llama-3.2-1B & 10 & Docling & 42.0 & 64.6 & 53.2 & 1.5 & 2.6 \\
Llama-3.2-1B & 10 & Tesseract & 42.9 & 65.2 & 53.9 & 4.4 & 5.6 \\
Llama-3.2-3B & 10 & DeepSeek & 56.5 & 74.9 & 66.2 & 75.1 & 79.3 \\
Llama-3.2-3B & 10 & Docling & 59.8 & 78.6 & 68.7 & 1.5 & 4.4 \\
Llama-3.2-3B & 10 & Tesseract & 59.4 & 77.2 & 68.6 & 4.4 & 7.4 \\
Ministral-3-3B & 10 & DeepSeek & 71.2 & 87.5 & 79.3 & 75.1 & 79.0 \\
Ministral-3-3B & 10 & Docling & 73.2 & 89.2 & 80.3 & 1.5 & 4.0 \\
Ministral-3-3B & 10 & Tesseract & 75.9 & 90.4 & 82.5 & 4.4 & 6.9 \\
Mistral-7B & 10 & DeepSeek & 72.3 & 89.0 & 79.1 & 75.1 & 85.6 \\
Mistral-7B & 10 & Docling & 70.9 & 88.1 & 78.3 & 1.5 & 8.2 \\
Mistral-7B & 10 & Tesseract & 71.0 & 87.3 & 79.0 & 4.4 & 11.3 \\
NuExtract-2.0-4B & 1 & - & 52.1 & 65.5 & 59.6 & - & 11.4 \\
Qwen3-0.6B & 10 & DeepSeek & 57.6 & 78.4 & 66.2 & 75.1 & 76.7 \\
Qwen3-0.6B & 10 & Docling & 56.0 & 77.5 & 65.1 & 1.5 & 2.4 \\
Qwen3-0.6B & 10 & Tesseract & 61.3 & 79.6 & 69.5 & 4.4 & 5.4 \\
Qwen3-1.7B & 10 & DeepSeek & 72.3 & 90.2 & 79.3 & 75.1 & 77.9 \\
Qwen3-1.7B & 10 & Docling & 71.6 & 89.7 & 78.6 & 1.5 & 3.3 \\
Qwen3-1.7B & 10 & Tesseract & 73.1 & 89.9 & 80.2 & 4.4 & 6.2 \\
Qwen3-4B & 10 & DeepSeek & 74.7 & 91.9 & 82.3 & 75.1 & 80.8 \\
Qwen3-4B & 10 & Docling & 75.4 & 92.4 & 82.4 & 1.5 & 5.0 \\
Qwen3-4B & 10 & Tesseract & 76.9 & 93.0 & 84.1 & 4.4 & 7.9 \\
Qwen3-8B & 10 & DeepSeek & 71.9 & 91.0 & 80.1 & 75.1 & 84.5 \\
Qwen3-8B & 10 & Docling & 71.2 & 90.2 & 79.0 & 1.5 & 7.4 \\
Qwen3-8B & 10 & Tesseract & 74.4 & 92.1 & 81.9 & 4.4 & 10.3 \\
Qwen3-VL-2B & 10 & - & 57.1 & 78.3 & 66.5 & - & 5.1 \\
Qwen3-VL-4B & 10 & - & 70.1 & 91.1 & 78.5 & - & 8.1 \\
Qwen3-VL-8B & 10 & - & 66.6 & 89.7 & 76.0 & - & 26.8 \\
\bottomrule
\end{tabular}

\vspace{2pt}
{\footnotesize $^\dagger$ In-domain: fine-tuned on Kleister-NDA by its authors; not comparable to the zero-shot rows.}
\end{table*}

\end{document}